\documentclass[10pt,twocolumn,letterpaper]{article}

\usepackage{cvpr}
\usepackage{times}
\usepackage{epsfig}
\usepackage{graphicx}
\usepackage{amsmath}
\usepackage{amssymb}
\usepackage{amsmath}
\usepackage{graphicx}
\usepackage{float}
\usepackage{listings}
\usepackage{subcaption}
\usepackage{algorithm}
\usepackage{algpseudocode}
\usepackage{fancyhdr}
\usepackage[titletoc]{appendix}
\usepackage{array}
\usepackage{booktabs}
\DeclareMathOperator*{\argmax}{\arg\!\max}

\usepackage[pagebackref=true,breaklinks=true,letterpaper=true,colorlinks,bookmarks=false]{hyperref}

\cvprfinalcopy 

\def\cvprPaperID{****} 

\ifcvprfinal\fi
\begin{document}
	
	\title{Efficient Image Evidence Analysis of CNN  Classification Results}
	
	\author{Keyang Zhou 
		and
		Bernhard Kainz\\
		Imperial College London\\
		180 Queen's Gate\\
		{\tt\small keyang.zhou16@imperial.ac.uk; bkainz@imperial.ac.uk}
	}
	
	\maketitle
	
	\begin{abstract}
	Convolutional neural networks (CNNs) define the current state-of-the-art for image recognition. With their emerging popularity, especially for critical applications like medical image analysis or self-driving cars, confirmability is becoming an issue.  The black-box nature of trained predictors make it difficult to trace failure cases or to understand the internal reasoning processes leading to results.
	In this paper we introduce a novel efficient method to visualise evidence that lead to decisions in CNNs. In contrast to network fixation or saliency map methods, our method is able to illustrate the evidence for or against a classifier's decision in input pixel space approximately $10/times$ faster than previous methods. 	
	 We also show that our approach is less prone to noise and can focus on the most relevant input regions, thus making it more accurate and interpretable. Moreover, by making simplifications we link our method with other visualisation methods, providing a general explanation for gradient-based visualisation techniques. We believe that our work makes network introspection more feasible for debugging and understanding deep convolutional networks. This will increase trust between humans and deep learning models.
	\end{abstract}
	
	\section{Introduction}
	
	CNNs learn to construct a hierarchy of representations from training images. These representations are incomprehensible to humans. It is challenging to gain a good understanding of what a trained network has learned without using appropriate visualisation techniques. There have been several efforts devoted to visualising deep neural network inference in recent years. These techniques complement each other to show different aspects of a CNN inference.
	An early approach to network introspection is based on Activation Maximization. 
	\emph{Activation Maximization} aims to find out what a neuron has learned by finding the image to which a neuron is maximally activated. Then the pattern in the image can be thought as an approximation to the learned representation.
	Activation Maximization collects sample images from the training set that have the highest activations for the neuron. By comparing them for similarities a common pattern can be found. This na\"ive approach is easy to implement but has several drawbacks. First, there is no quantitative measure of similarity and human observation is ambiguous and error-prone. Images may be interpreted in several ways. 
	Second, the whole training set has to be fed into the model to compute activation of neurons for every image, which is very time-consuming. Hence, it is often preferred to synthesize an image that can maximally activate a given neuron.

	The aim is to maximize the activation of a neuron, thus this can be treated as an optimisation problem \cite{erhan2009visualizing}. Let ${a_i}\left( {x;\theta } \right)$ be the activation of neuron $i$. It is a function of input image $x$ and model parameters $\theta$. We want to find a specific input $x^*$ that maximizes $a_i$ with 
	\begin{align}
	{x^*} = \argmax_{\left\| x \right\| = p} {a_i}\left( {x;\theta } \right).
	\end{align}
	This optimisation problem can be solved by gradient ascent. To start with, $x$ is an image filled with random pixels. In each iteration, the activation $a_i$ is computed. Its derivative with respect to $x$ is then used to update the pixel intensity of $x$. This process is repeated until convergence, and the resulting image should have high activation for neuron $i$.

	Applying this technique to neurons in hidden layers can produce some recognisable features learned by the network. However, if the aim is to visualise neurons in the output layer, this procedure would fail and generate images interpretable to humans \cite{nguyen2015deep}. An improvement was made by Simonyan et al. \cite{simonyan2013deep}. They added L2 regularization to the gradient ascent formulation and proposed 
	\begin{align}
	{x^*} = \argmax_{x} \left({a_i}\left( {x;\theta } \right) - \lambda \left\| x \right\|_2^2\right).
	\label{eq:Simonyan}
	\end{align}
	In Eq.~\ref{eq:Simonyan}, $\lambda$ is a regularization parameter. It controls the degree to which $x$ is penalized. With this small change in the objective function, gradient ascent generates more interpretable images for neurons in the output layer.

\begin{figure}[htb]
		\centering
		\includegraphics[width=0.8\columnwidth]{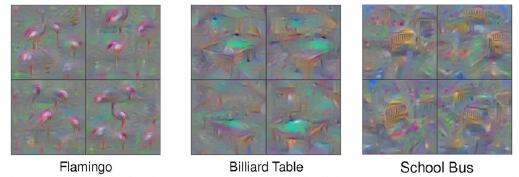}
		\caption{visualisation of output layer features \cite{yosinski2015understanding}.}
		\label{actmax}
	\end{figure}

\begin{figure}[htb]
		\includegraphics[width=\columnwidth]{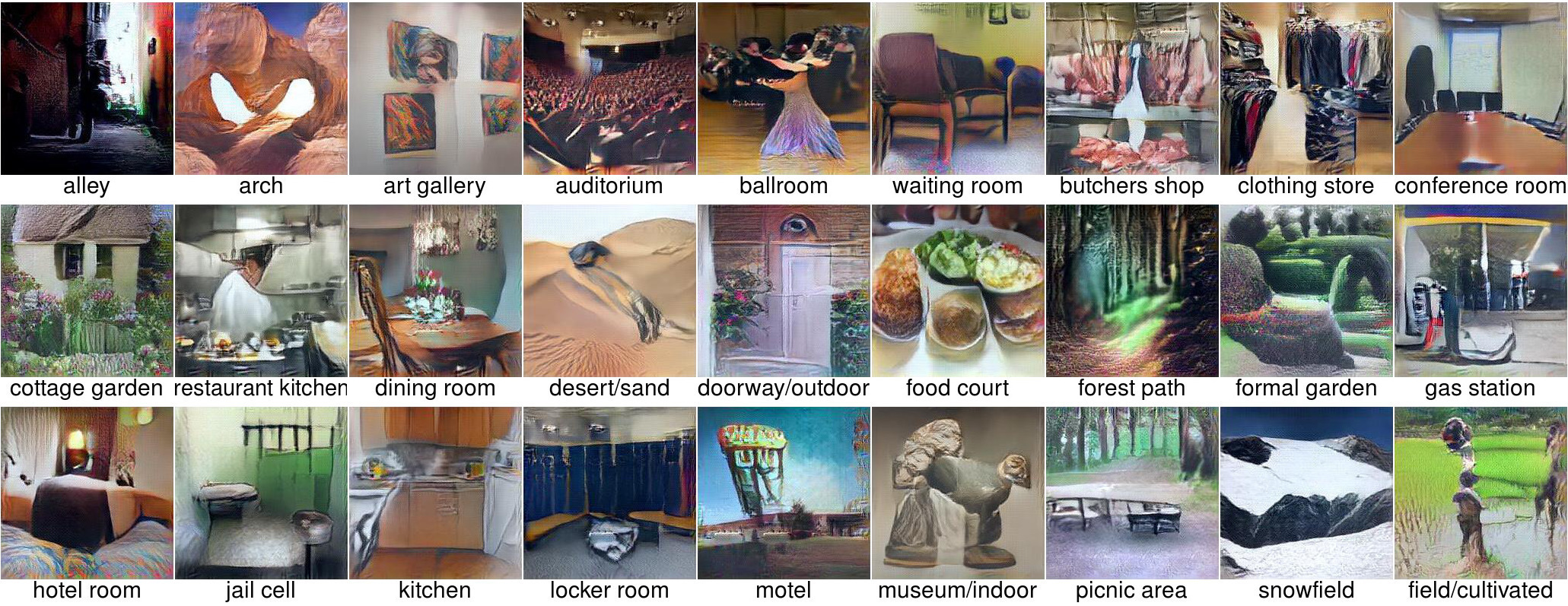}
		\caption{synthesized images that maximally activate given output neurons \cite{nguyen2016synthesizing}.}
		\label{synthesize}
	\end{figure}

	Based on this work, Yosinski et al. \cite{yosinski2015understanding} further generalized the gradient ascent updating rule. Instead of sticking to L2 regularization, they regularized an operator $r_{\theta}$ and experimented with different options as  
	\begin{align}
	x \leftarrow {r_\theta }\left( {x + \eta \frac{{\partial {a_i}}}{{\partial x}}} \right).
	\end{align}
	Here, $r_{\theta}$ may refer to L2 regularization, Gaussian blur or pixel clipping. In practice, a combination of these regularizers has shown to produce the most natural images. Figure \ref{actmax} shows the visualization of output layer neurons in an 8-layer CNN using this approach. 
	\noindent
	The images in Figure~\ref{actmax} can capture class-specific features to some extent, but they are too abstract and noisy for human observers to infer useful information. A different approach of visualisation synthesis was adopted by Nguyen et al. \cite{nguyen2016synthesizing}. Instead of performing gradient ascent directly on the input space of the model, they use a generator network $G$ for generating visualisations and optimise on the input space of $G$ by
	\begin{align}
	{x^*} = \argmax_{x}\left( {{a_i}\left( {G\left( x \right)} \right) - \lambda \left\| x \right\|_2^2} \right).
	\end{align}
	Since $G$ was explicitly trained to generate natural images, it acts as a prior to ensure the interpretability of visualizations. Example images are shown in Figure \ref{synthesize}.
	\noindent

	Activation maximization is helpful for human observers to understand the general representation learned by a deep neural network. It cannot explain the network's response to an individual image.  \emph{Representation Inversion} has been developed as a class of techniques tailored for this purpose. 
	It works by projecting the output or hidden activations of a network back to input space and visualising the result. 

	A classic Representation Inversion algorithm is deconvolution. As its name suggests, a deconvolutional network (DCNN) performs the inverse operations of a CNNs \cite{zeiler2011adaptive}. In correspondence with a CNN, a DCNN has three components: unpooling, rectification and filtering.
	To visualise a given neuron, a DCNN is attached to the layers of a CNN and all other neuron activations are set to zero. Feeding the feature map as input to the DCNN, the inverse of all operations in the CNN are performed in reverse order until input space is reached.
	\begin{figure}[htb]
		\centering
		\includegraphics[width=\columnwidth]{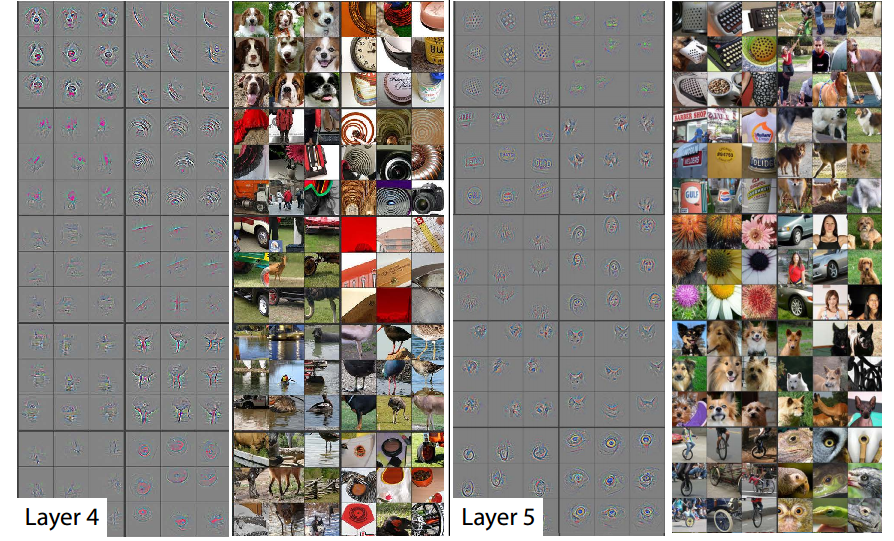}
		\caption{image patches and the corresponding reconstructed patterns \cite{zeiler2014visualizing}.}
		\label{deconvolution}
	\end{figure}
	\noindent
	Figure \ref{deconvolution} shows the visualisation of high-level features of a trained CNN. We can observe that patterns reconstructed via deconvolution have similar shapes for images in the same class despite their individual differences. This might not be desirable if we want to visualise more image-specific features. 

	Simonyan et al. proposed to provide class saliency maps by computing partial derivative of  the class score with respect to each input pixel \cite{simonyan2013deep}. This can be thought as a Representation Inversion method because gradients information is propagated back in this case.
	\begin{figure}[htb]
		\centering
		\includegraphics[width=0.8\columnwidth]{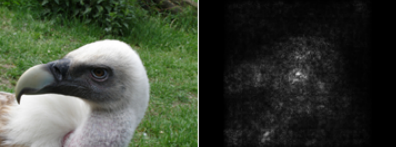}
		\caption{an image and its corresponding saliency map \cite{simonyan2013deep}.}
		\label{sensitivity}
	\end{figure}

	Figure \ref{sensitivity} demonstrates an example of a class saliency map. The object's contour is roughly preserved in the visualisation. This method is fast to compute, but it only illustrates the sensitivity of a model's prediction to individual pixels. It cannot be used to accurately measure each pixel's relevance since it does not consider higher-order interactions of pixels.

	Bach et al. proposed the principle of conservation for pixel relevance distribution; the sum of relevance of pixel $i$, $R_i$, should be roughly equal to the model's output $f(x)$ \cite{bach2015pixel}. Following this principle, a distribution rule called layer-wise relevance propagation is given by
	\begin{align}
	R_i^l = \sum\limits_j {\frac{{{z_{ij}}}}{{\sum\limits_{i'} {{z_{i'j}}} }}R_j^{l + 1}},
	\end{align}
	where $R_i^l$ is relevance of neuron $i$ in layer $l$ and $z_{ij}$ equals $x_i^lw_{ij}^{l + 1}$.

	This rule assigns relevance of upper-layer neurons to a lower-layer neuron proportionally to their connecting weights. Since there is a non-linear activation function between layers, an approximation rule (deep Taylor decomposition) is used \cite{montavon2017explaining}, which is given by
	\begin{align}
	{R_i} = {\frac{{\partial f}}{{\partial {{\tilde x}_i}}} \cdot \left( {{x_i} - {{\tilde x}_i}} \right)},
	\end{align}
	where $\tilde x$ is chosen such that $f\left(\tilde x\right) = 0$.
	\begin{figure}[htb]
		\centering
		\includegraphics[width=0.8\columnwidth]{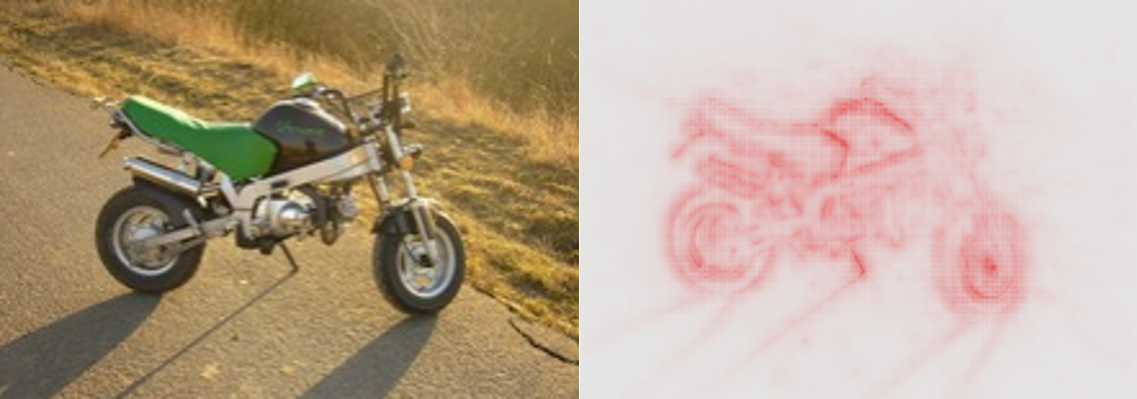}
		\caption{an image and its corresponding deep Taylor decomposition \cite{montavon2017explaining}}
		\label{taylor}
	\end{figure}
	\noindent
	Selvaraju et al. used another approach to distribute feature relevance: \emph{Gradient-weighted Class Activation Mapping} (Grad-CAM) \cite{selvaraju2016grad} is a generalization of \emph{Class Activation Mapping} (CAM) \cite{zhou2016learning}. It is based on the observations that deeper layers of a CNN usually encode higher-level visual representations, and spatial locations are lost in fully connected layers. So, the last convolutional layer preserves both semantic and spatial information.

\begin{figure}[htb]
		\centering
		\includegraphics[width=\columnwidth]{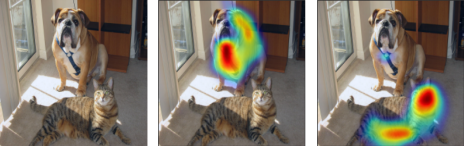}
		\caption{an image and its corresponding Grad-CAM for different classes \cite{selvaraju2016grad}}
		\label{gradcam}
	\end{figure}

	Grad-CAM computes first the partial derivatives of class scores with respect to each feature map $A^k$ of the last convolutional layer and takes the average to get feature map importance $a_k$:
	\begin{align}
	{a_k} = \frac{1}{Z}\sum\limits_i {\sum\limits_j {\frac{{\partial y}}{{\partial A_{ij}^k}}} }, 
	\end{align}
	where $Z$ is feature map size. Then the class activation map is computed by
	\begin{align}
	{L_{CAM}} = {\mathop{ReLU}}\left( {\sum\limits_k {{a_k}{A^K}} } \right) \odot \frac{{\partial f}}{{\partial x}}.
	\end{align}
	Since $L_{CAM}$ is of the same size as the last feature map, it needs to be up-sampled to input size for visualisation, as shown in Figure \ref{gradcam}.

	An alternative way to visualise the decision processes of black-box predictors is through occluding parts of an input image and observe the changes in a predictor's performance. If the prediction changes significantly after occluding a patch, pixels in that patch are assigned higher importance. Zeiler et al. used grey patches for occlusion and produced heatmaps to show evidence for and against a classification decision \cite{zeiler2014visualizing}.

	Zintgraf et al. argued in their work that Zeiler's approach is inaccurate because grey patches would feed new information into the model \cite{zintgraf2017visualizing}. Instead, they applied \emph{Prediction Difference Analysis} (PDA) and measured a pixel's importance by the change in a classifier's output after marginalizing out that pixel. The marginalized classifier output is derived from Bayes’ rule: 
	\begin{align}
	P\left( {c|{x_{\backslash i}}} \right) = \sum\limits_{{x_i}} {P\left( {{x_i}|{x_{\backslash i}}} \right)P\left( {c|{x_i},{x_{\backslash i}}} \right)}, 
	\label{eq:PDA}
	\end{align}
	where $c$ is predicted class of the input image, $x_i$ is the $i$-th pixel and $x_{\backslash i}$ denotes all pixels of the image except $x_i$.

	A few changes have been made to accommodate this technique to deep neural networks. First, instead of a single pixel, a small patch of pixels with size $k \times k$ is marginalized out each time for larger fluctuations in model prediction. Second, for images of large size $n \times n$, sampling a window of pixels with multivariate Gaussian distribution conditioned on the remaining pixels is clearly infeasible. So, a patch surrounding the window with size $l \times l$ is chosen so that sampling is conditioned on that outer patch. The algorithm is demonstrated in pseudo-code in Algorithm~\ref{pda_code}.
	
	\begin{algorithm}
		\caption{Prediction Difference Analysis \cite{zintgraf2017visualizing}}
		\label{pda_code}
		\begin{algorithmic}[1]
			\State $WE =$ zeros($n*n$), $counts =$ zeros($n*n$)
			\For {every patch $x_w$ of size $k \times k$ in $x$}
			\State $x'$ = copy($x$)
			\State $sum_w = 0$
			\State define patch ${\hat x_w}$ of size $l \times l$ that contains $x_w$
			\For{$s=1$ \textbf{to} $S$}
			\State ${x_w'} = {x_w}$ sampled from $P\left( {\left. {{x_w}} \right|{{\mathord{\buildrel{\lower3pt\hbox{$\scriptscriptstyle\frown$}} \over x} }_w}\backslash {x_w}} \right)$
			\State $sum_w$  += $p(x')$
			\EndFor
			\State $P\left( {\left. c \right|x\backslash {x_w}} \right) = sum_w / S$
			\State $WE[$coordinates of $x_w]$ += ${\log _2}\left( {odds\left( {c|x} \right)} \right) - {\log _2}\left( {odds\left( {c|x\backslash {x_w}} \right)} \right)$
			\State $counts[$coordinates of $x_w]$ += $1$
			\EndFor
			\State \textbf{return} $WE / counts$
		\end{algorithmic}
	\end{algorithm}

	\begin{figure}[htb]
		\centering
		\includegraphics[width=0.8\columnwidth]{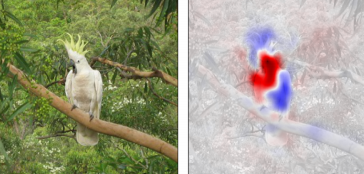}
		\caption{an image and its corresponding prediction difference analysis \cite{zintgraf2017visualizing}.}
		\label{pda}
	\end{figure}
	\noindent
	Figure \ref{pda} shows an example of PDA. In the visualisation, red regions show evidence supporting a model's prediction, while blue regions show evidence against a model's prediction.

	This technique has several limitations. First, it is very slow because of tens of thousands of Gaussian conditional sampling and deep network forward passes. It could take more than an hour to visualise a single image for some large models. Moreover, the authors approximated Eq.~\ref{eq:PDA} by taking only ten samples for computational reasons, which is a source of error. Lastly, a Gaussian distribution conditioned on local patch does not take global context into account, making the conditional probability approximation less accurate because local patches have been shown to have different semantics under different contexts \cite{rabinovich2007objects}. Although easy to compute, Gaussian distribution itself is not a good model for natural images \cite{ruderman1994statistics}.
	
	In this paper we propose an alternative formulation of PDA to make this method \textbf{(1)} efficient enough to be applicable for the practice, \textbf{(2)} more accurate, thus providing better interpretability, and \textbf{(3)} link it to other gradient-based visualisation techniques. Our approach is up to $10\times$ faster than the original formulation and provides a comprehensive mathematical framework for such approaches.  
 		
\section{Method}
	
In the following we will first make an informal analysis of the complexity of PDA. This can help us intuitively understand where the major performance bottlenecks are. Then, an alternative formulation for this algorithm will be designed to avoid these bottlenecks while producing visualisations with similar quality.

We shall also discuss the circumstances under which this formulation would give accurate result, and in turn show that it holds for most of modern CNN architectures. In Section~\ref{sect:eval}, we will run benchmark experiments on various classifiers to quantitatively compare the runtime of the original version of PDA and the efficient version proposed by us.

\textbf{Complexity of PDA:} Algorithm~\ref{pda_code} mainly consists of an outer loop and an inner loop. The outer loop iterates over every patch in the input image, and the inner loop takes $S$ samples for each given patch and performs an equal number of forward passes of the classifier to get the average prediction.

Suppose the input image is of size $n \times n$, a patch is of size $k \times k$ and the outer patch it conditioned on is of size $l \times l$. Then there are $ \left( {n - k + 1} \right) \times \left( {n - k + 1} \right)$ patches in total in the image. Hence, there are $ S\left( {n - k + 1} \right)^2$ samples to be taken. If we allow the classifier to operate on small batches of size $m$, then it would run $\frac{{S{{\left( {n - k + 1} \right)}^2}}}{m}$ forward passes.

Hence, the performance of PDA depends on sampling time and forward pass time.

\textbf{Sampling Time:} One common way to draw a sample $y$ from a multivariate Gaussian distribution is through Cholesky decomposition, which takes the form $\Sigma  = L{L^T}$. If $\Sigma$ is positive definite, this decomposition is unique. Suppose the multivariate Gaussian distribution, which we want to sample from, has mean $\mu$ and covariance matrix $\Sigma$. We could first decompose $\Sigma$ to get $L$. Then, we draw a sample $x$ from a standard multivariate Gaussian distribution with mean $0$ and identity covariance matrix $I$, and transform it by $y = Lx + \mu$.

Finding the exact time complexity of multivariate Gaussian distribution sampling requires to first analyse the complexity of Cholesky decomposition and standard Gaussian distribution sampling. However, since the practical performance of these basic matrix operations strongly depend on the low level BLAS implementation and CPU instruction set, we will empirically measure the sampling time instead of focusing on theory.

Let the image size be $224 \times 224$, patch size be $10 \times 10$ and number of samples taken for each patch be $10$, which are the settings adopted by Zintgraf et al. \cite{zintgraf2017visualizing} in their  implementation of PDA. Hence there are $224 \times 224 \times 10 = 501760$ samples of $100$ dimensions to be drawn.  Under Intel Core i5 CPU and NumPy\footnote{the fundamental scientific computing library for Python: \url{www.numpy.org}}, it is measured that the sampling process takes roughly $5$ minutes on average, which is a considerable cost.

\textbf{Forward Pass Time:} Since the inference speed differs significantly for different CNN architectures, we will again run experiments to empirically measure the time spent on forward pass. We will use a batch size of $160$ on a Tesla K80 GPU. Caffe\footnote{a deep learning framework specialized in vision applications: \url{caffe.berkeleyvision.org}} \cite{jia2014caffe} with CuDNN\footnote{a library of primitives for deep learning with CUDA support: \url{developer.nvidia.com/cudnn}} \cite{chetlur2014cudnn} support is used in the following experiments. All models used here are pretrained and available from the Caffe Model Zoo.

Among various CNN architectures we choose to measure inference speed of AlexNet, VGG-16 and GoogLenet for two reasons. First, these architectures were winners of past ILSVRC classification task, and they have remained influential since many other models are finetuned with respect to them. Also, it would allow direct comparisons with PDA since these architectures were also used by Zintgraf et al. for their experiments.

\begin{table}[htb]
	\begin{center}
		\begin{tabular}{| c | c | c |}
			\hline
			Architecture & Input Size & Forward Pass Time (ms) \\ \hline
			AlexNet & $227 \times 227$ & 342 \\ \hline
			VGG-16 & $224 \times 224$ & 3301 \\ \hline
			GoogLenet & $224 \times 224$ & 634 \\ \hline
		\end{tabular}
		\caption{Inference time for three popular CNN architectures.}
		\label{benchmark}
	\end{center}
\end{table}
\noindent
Table~\ref{benchmark} summarizes the single batch forward pass time for AlexNet, VGG-16 and GoogLeNet respectively. VGG-16 is especially computationally expensive in inference. It takes more than $3$ seconds to compute predictions for a batch of $160$ input images.  Since $501760$ samples can be put into $3136$ batches exactly, it would take a VGG-16 model approximately $172$ minutes to do all the inference work in order to visualize PDA for a single image, which is too long to be useful in real scenarios.

Note that there are other operations which we have not discussed, such as fitting conditional Gaussian distributions for each patch. However, these distribution parameters can be computed in advance and loading them would only incur minimal computational overhead.

\subsection{Alternative Formulation of PDA}
From the empirical results above, it is obvious that we cannot achieve a significant speedup without reducing the number of samples to be taken and the number of forward passes to be run.

We observe that in PDA the class probability after marginalizing out a small window of pixels $P\left( {\left. c \right|x\backslash {x_w}} \right)$ is approximated by
\begin{align}
\sum\limits_{{x_w}} {P\left( {{x_w}|{{\mathord{\buildrel{\lower3pt\hbox{$\scriptscriptstyle\frown$}} 
					\over x} }_w}\backslash {x_w}} \right)P\left( {c|{x_w},x\backslash {x_w}} \right)}, 
\end{align}
which is the arithmetic mean of ${P\left( {c|{x_w},x\backslash {x_w}} \right)}$. If we substitute this with the geometric mean, we will get $\prod\limits_{{x_w}} {P{{\left( {c|{x_w},x\backslash {x_w}} \right)}^{P\left( {{x_w}|{{\mathord{\buildrel{\lower3pt\hbox{$\scriptscriptstyle\frown$}} \over x} }_w}\backslash {x_w}} \right)}}} $. Since the geometric mean of a probability distribution is not necessarily itself a probability distribution, we need to first normalize it:
\begin{align}
\begin{split}
P\left( {\left. c \right|x\backslash {x_w}} \right) \approx \frac{{\prod\limits_{{x_w}} {P{{\left( {c|{x_w},x\backslash {x_w}} \right)}^{P\left( {{x_w}|{{\mathord{\buildrel{\lower3pt\hbox{$\scriptscriptstyle\frown$}} 
									\over x} }_w}\backslash {x_w}} \right)}}} }}{{\sum\limits_c {\prod\limits_{{x_w}} {P{{\left( {c|{x_w},x\backslash {x_w}} \right)}^{P\left( {{x_w}|{{\mathord{\buildrel{\lower3pt\hbox{$\scriptscriptstyle\frown$}} 
										\over x} }_w}\backslash {x_w}} \right)}}} } }}.
\end{split}
\end{align}
For a CNN, the last layer is often a softmax layer:
\begin{align}
\begin{split}
P\left( {\left. c \right|x} \right) &= softmax {\left( {{z^l}} \right)_c}\\
&= \frac{{\exp \left( {z_c^l} \right)}}{{\sum\limits_j {\exp \left( {z_j^l} \right)} }},
\end{split}
\label{eq:oldpda}
\end{align}
where $z^l$ is the last layer before softmax and $z_j^l$ is its $j$-th neuron. So, 
\begin{align}
\begin{split}
&P\left( {\left. c \right|x\backslash {x_w}} \right) \approx \\
&\approx \frac{{\prod\limits_{{x_w}} {softmax \left( {{z^l}} \right)_c^{P\left( {{x_w}|{{\mathord{\buildrel{\lower3pt\hbox{$\scriptscriptstyle\frown$}} 
								\over x} }_w}\backslash {x_w}} \right)}} }}{{\sum\limits_j {\prod\limits_{{x_w}} {softmax \left( {{z^l}} \right)_j^{P\left( {{x_w}|{{\mathord{\buildrel{\lower3pt\hbox{$\scriptscriptstyle\frown$}} 
									\over x} }_w}\backslash {x_w}} \right)}} } }}\\
&= \frac{1}{{\sum\limits_j {\prod\limits_{{x_w}} {{{\left( {\frac{{softmax {{\left( {{z^l}} \right)}_j}}}{{softmax {{\left( {{z^l}} \right)}_c}}}} \right)}^{P\left( {{x_w}|{{\mathord{\buildrel{\lower3pt\hbox{$\scriptscriptstyle\frown$}} \over x} }_w}\backslash {x_w}} \right)}}} } }}\\
&= \frac{1}{{\sum\limits_j {\prod\limits_{{x_w}} {{{\left( {\frac{{\exp \left( {z_j^l} \right)}}{{\exp \left( {z_c^l} \right)}}} \right)}^{P\left( {{x_w}|{{\mathord{\buildrel{\lower3pt\hbox{$\scriptscriptstyle\frown$}} \over x} }_w}\backslash {x_w}} \right)}}} } }}\\
&= \frac{1}{{\sum\limits_j {\prod\limits_{{x_w}} {\exp {{\left( {z_j^l - z_c^l} \right)}^{P\left( {{x_w}|{{\mathord{\buildrel{\lower3pt\hbox{$\scriptscriptstyle\frown$}} 
										\over x} }_w}\backslash {x_w}} \right)}}} } }}\\
&= \frac{1}{{\sum\limits_j {\exp \left( {\sum\limits_{{x_w}} {P\left( {{x_w}|{{\mathord{\buildrel{\lower3pt\hbox{$\scriptscriptstyle\frown$}} 
									\over x} }_w}\backslash {x_w}} \right)\left( {z_j^l - z_c^l} \right)} } \right)} }}\\
&= \frac{1}{{\sum\limits_j {\frac{{\exp \left( {\sum\limits_{{x_w}} {P\left( {{x_w}|{{\mathord{\buildrel{\lower3pt\hbox{$\scriptscriptstyle\frown$}} 
											\over x} }_w}\backslash {x_w}} \right)z_j^l} } \right)}}{{\exp \left( {\sum\limits_{{x_w}} {P\left( {{x_w}|{{\mathord{\buildrel{\lower3pt\hbox{$\scriptscriptstyle\frown$}} 
											\over x} }_w}\backslash {x_w}} \right)z_c^l} } \right)}}} }}\\
\end{split}
\label{eq:extpda1}
\end{align}
\begin{align}
\begin{split}
&= \frac{{\exp \left( {\sum\limits_{{x_w}} {P\left( {{x_w}|{{\mathord{\buildrel{\lower3pt\hbox{$\scriptscriptstyle\frown$}} 
								\over x} }_w}\backslash {x_w}} \right)z_c^l} } \right)}}{{\sum\limits_j {\exp \left( {\sum\limits_{{x_w}} {P\left( {{x_w}|{{\mathord{\buildrel{\lower3pt\hbox{$\scriptscriptstyle\frown$}} 
									\over x} }_w}\backslash {x_w}} \right)z_j^l} } \right)} }}\\
&= softmax {\left( {\sum\limits_{{x_w}} {P\left( {{x_w}|{{\mathord{\buildrel{\lower3pt\hbox{$\scriptscriptstyle\frown$}} 
							\over x} }_w}\backslash {x_w}} \right){z^l}} } \right)_c}
\end{split}
\label{eq:extpda}
\end{align}
Comparing Eq~\ref{eq:oldpda} to Eq~\ref{eq:extpda}, we can immediately observe that $P\left( {\left. c \right|x\backslash {x_w}} \right)$ is roughly equal to applying the softmax function to the conditional expectation of $z^l$, ${\rm E}\left[ {{z^l}|{{\mathord{\buildrel{\lower3pt\hbox{$\scriptscriptstyle\frown$}} \over x} }_w}\backslash {x_w}} \right]$.

What we have just shown is that the conditional probability ${P\left( {{x_w}|{{\mathord{\buildrel{\lower3pt\hbox{$\scriptscriptstyle\frown$}} \over x} }_w}\backslash {x_w}} \right)}$ can be pushed into a convolutional network's decision function if we approximate the arithmetic mean by the corresponding normalized geometric mean. This observation leads to 
\begin{align}
\begin{split}
&P\left( {\left. c \right|x\backslash {x_w}} \right)\approx\\
&\approx \sum\limits_{{x_w}} {P\left( {{x_w}|{{\mathord{\buildrel{\lower3pt\hbox{$\scriptscriptstyle\frown$}} 
					\over x} }_w}\backslash {x_w}} \right)softmax {{\left( {{z^l}\left( {{x_w},x\backslash {x_w}} \right)} \right)}_c}} \approx\\
&\approx softmax {\left( {\sum\limits_{{x_w}} {P\left( {{x_w}|{{\mathord{\buildrel{\lower3pt\hbox{$\scriptscriptstyle\frown$}} 
							\over x} }_w}\backslash {x_w}} \right){z^l}\left( {{x_w},x\backslash {x_w}} \right)} } \right)_c}.
\end{split}
\end{align}
The arithmetic mean is now taken in layer $l$, the last layer before softmax. It can reduce the number of softmax functions to be evaluated for each patch from $S$ to $1$. This is not so satisfactory because we still have to run $S$ forward passes for each patch, except that each forward pass stops at layer $l$.

As explained previously, a CNN is composed of many layers of different types, and each layer can be thought of as a simple function in this composition of functions. So, to gain further speed-up we need to investigate whether the arithmetic mean could be taken at lower layers. Specifically, we are looking for component functions $f$ of a convolutional network that satisfies either 
\begin{align}
\rm E\left[ {f\left( x \right)} \right] = f\left( {{\rm E}\left[ x \right]} \right)
\label{eg:E}
\end{align}
or
\begin{align}
GM\left( {f\left( x \right)} \right) = f\left( {{\rm E}\left[ x \right]} \right),
\label{eg:GM}
\end{align}
where $GM$ denotes (normalized) geometric mean.

If $f$ satisfies Eq.~\ref{eg:E}, the arithmetic mean can be propagated to a lower layer exactly. If $f$ satisfies Eq.~\ref{eg:GM}, we can use geometric mean for $f$ as an approximation in the same way as softmax. In the following, we will investigate linear components, piece-wise linear components and non-linear components of a CNN respectively and prove that they possess the required properties.

\textbf{Linear Components:} Linear transformations constitute an important part of many classifiers because they allow more efficient optimisation and inference. The most common linear transformations in a CNN include convolutional layers, fully connected layers and batch normalization layers.

By linearity of expectation, Eq.~\ref{eg:E} trivially holds for all linear transformations, which means the expectation can be propagated down through these layers without losing accuracy.

\textbf{Piece-wise Linear Components:} Piece-wise linear functions are often used in CNNs as activation functions to solve the vanishing gradient problem. Examples of this kind include rectified linear units and maxout units \cite{goodfellow2013maxout}. Also, the max pooling layer following a convolutional layer is piece-wise linear.

We first consider the case of a ReLU function. Assume the input $x$ is Gaussian distributed with mean $\mu$ and variance ${\sigma ^2}$. This assumption may not hold in reality, but it can give us some intuition on ReLU's property. Baldi et al. \cite{baldi2014dropout} prove that if $\mu = 0$,
\begin{align}
\left| {{\rm E}\left[ {{\mathop{\rm ReLU}\nolimits} \left( x \right)} \right] - {\mathop{\rm ReLU}\nolimits} \left( {{\rm E}\left[ x \right]} \right)} \right| = \frac{\sigma }{{\sqrt {2\pi } }}.
\label{eq:relu1}
\end{align}
Moreover, if $\frac{{\left| \mu  \right|}}{\sigma }$ is large,
\begin{align}
\left| {{\rm E}\left[ {{\mathop{\rm ReLU}\nolimits} \left( x \right)} \right] - {\mathop{\rm ReLU}\nolimits} \left( {{\rm E}\left[ x \right]} \right)} \right| \approx 0.
\label{eq:relu2}
\end{align}
Eq.~\ref{eq:relu1} and Eq.~\ref{eq:relu2} together indicate that reducing the error of approximation depends on variance of input being small. We are going to show that the same conclusion applies to more general cases.

\begin{figure}[htb]
	\centering
	\includegraphics[width=1.0\columnwidth]{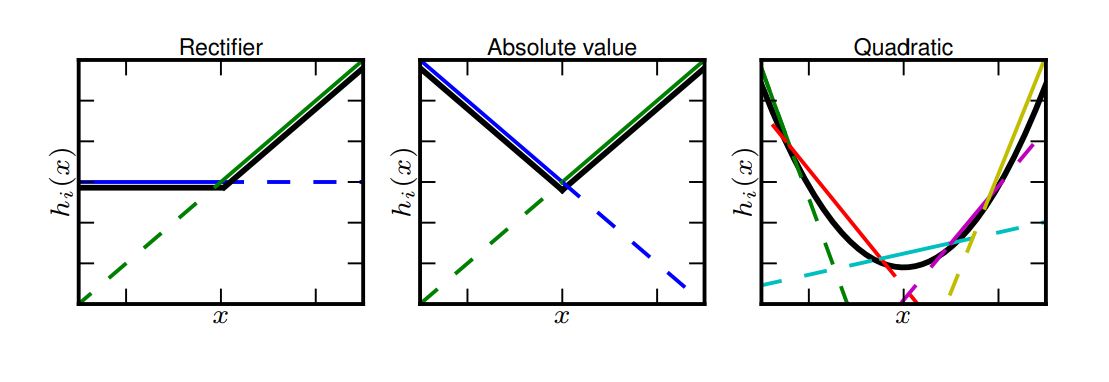}
	\caption{a maxout unit learns ReLU, absolute value function and approximates quadratic function \cite{goodfellow2013maxout}}
	\label{maxout}
\end{figure}

Specifically,\emph{maxout units} are capable of representing a broad class of piece-wise linear functions. A single maxout unit can approximate arbitrary convex functions. Figure~\ref{maxout} demonstrates how a maxout unit learns to behave like ReLU, absolute value function and quadratic function. Moreover, when applying to a convolutional layer, computing maxout activation is equivalent to performing max pooling across channels as well as spatial locations. So it covers the case of pooling layer as well.
\noindent
Let the input $x$ be a $d$-dimensional vector. A maxout unit $h(x)$ is defined by
\begin{align}
h\left( x \right) = \max \left( {{w_1}^Tx + {b_1},...,{w_k}^Tx + {b_k}} \right)
\end{align}
where $w_i$ and $b_i$ are parameters.

Assume elements of $x$ are independent and Gaussian distributed. Since the linear combination of independent Gaussian random variables remains Gaussian distributed, we can write
\begin{align}
h\left( x \right) = \max \left( {{{\tilde x}_1},...,{{\tilde x}_k}} \right), 
\end{align}
where ${{\tilde x}_i} = {w_i}^Tx + {b_i}$ and ${{\tilde x}_i} \sim N\left( {{{\tilde x}_i}|{\mu _i},\sigma _i^2} \right)$.

Since the maximum of a set of convex functions is convex, $h(x)$ is a convex function. Then by Jensen's inequality,
\begin{align}
{\rm E}\left[ {h\left( x \right)} \right] &= E\left[ {\max \left( {{{\tilde x}_1},...,{{\tilde x}_k}} \right)} \right]\\
& \ge \max \left( {{\rm E}\left[ {{{\tilde x}_1}} \right],..,{\rm E}\left[ {{{\tilde x}_k}} \right]} \right)\\
& = \max \left( {{\mu _1},...,{\mu _k}} \right).
\end{align}
On the other hand, again by Jensen's inequality
\begin{align}
\begin{split}
\exp \left( {t{\rm E}\left[ {h\left( x \right)} \right]} \right) &\le {\rm E}\left[ {\exp \left( {th\left( x \right)} \right)} \right]\\
& = {\rm E}\left[ {\exp \left( {t\max \left( {{{\tilde x}_1},...,{{\tilde x}_k}} \right)} \right)} \right]\\
& = {\rm E}\left[ {\max \left( {\exp \left( {t{{\tilde x}_1}} \right),...,\exp \left( {t{{\tilde x}_k}} \right)} \right)} \right]\\
& \le {\rm E}\left[ {\sum\limits_i {\exp \left( {t{{\tilde x}_i}} \right)} } \right]\\
& = \sum\limits_i {{\rm E}\left[ {\exp \left( {t{{\tilde x}_i}} \right)} \right]}
\end{split}
\end{align}
Since $\tilde x_i$ is Gaussian distributed, by definition of the moment generating function of Gaussian distributions,
\begin{align}
\exp \left( {t{{\tilde x}_i}} \right) = \exp \left( {{\mu _i}t + \frac{1}{2}{t^2}\sigma _i^2} \right).
\end{align}
Hence,
\begin{align}
\begin{split}
\exp \left( {t{\rm E}\left[ {h\left( x \right)} \right]} \right) &\le \sum\limits_i {{\rm E}\left[ {\exp \left( {t{{\tilde x}_i}} \right)} \right]} \\
& = \sum\limits_i {\exp \left( {{\mu _i}t + \frac{1}{2}{t^2}\sigma _i^2} \right)} 
\end{split}
\end{align}
Taking the logarithm on both sides, we get
\begin{align}
{\rm E}\left[ {h\left( x \right)} \right] \le 
\frac{1}{t}\log \sum\limits_i {\exp \left( {{\mu _i}t + \frac{1}{2}{t^2}\sigma _i^2} \right)}
\end{align}
We still need an upper bound for the log-sum-exp function, which can be derived by
\begin{align}
\begin{split}
\exp \left( {\log \sum\limits_i {\exp \left( {{x_i}} \right)} } \right) &= \sum\limits_i {\exp \left( {{x_i}} \right)} \\
& \le k\max \left( {\exp \left( {{x_1}} \right),...,\exp \left( {{x_2}} \right)} \right)\\
& = k\exp \left( {\max \left( {{x_1},...,{x_k}} \right)} \right)\\
\log \sum\limits_i {\exp \left( {{x_i}} \right)}  &\le \max \left( {{x_1},...,{x_k}} \right) + \log k
\end{split}
\end{align}
Without loss of generality we can set $t = 1$. We have the following bounds for ${\rm E}\left[ {h\left( x \right)} \right]$:
\begin{align}
\begin{split}
&\max \left( {{\mu _1},...,{\mu _k}} \right) \le \\
&{\rm E}\left[ {h\left( x \right)} \right] \le \log k + \max \left( {{\mu _1} + \frac{1}{2}\sigma _1^2,...,{\mu _k} + \frac{1}{2}\sigma _k^2} \right).
\end{split}
\end{align}
So the approximation error is
\begin{align}
\begin{split}
&\left| {{\rm E}\left[ {h\left( x \right)} \right] - h\left( {{\rm E}\left[ x \right]} \right)} \right|  \le 
\log k + \frac{1}{2}\max \left( {\sigma _1^2,...,\sigma _k^2} \right) + \\
& + \max \left( {{\mu _1},...,{\mu _k}} \right) - \max \left( {{\mu _1},...,{\mu _k}} \right) = \\
& = \log k + \frac{1}{2}\max \left( {\sigma _1^2,...,\sigma _k^2} \right)
\end{split}
\label{eq:aerror}
\end{align}
Eq.~\ref{eq:aerror} shows that the approximation quality for maxout units depends on the maximum of input variance.

We have shown that the error incurred by pushing the expectation inside a piece-wise linear function is small given that the variance of input is small. To find out whether it holds for a convolutional network, we design the following experiment.

\begin{figure}[htb]
	\centering
	\includegraphics[width=\columnwidth]{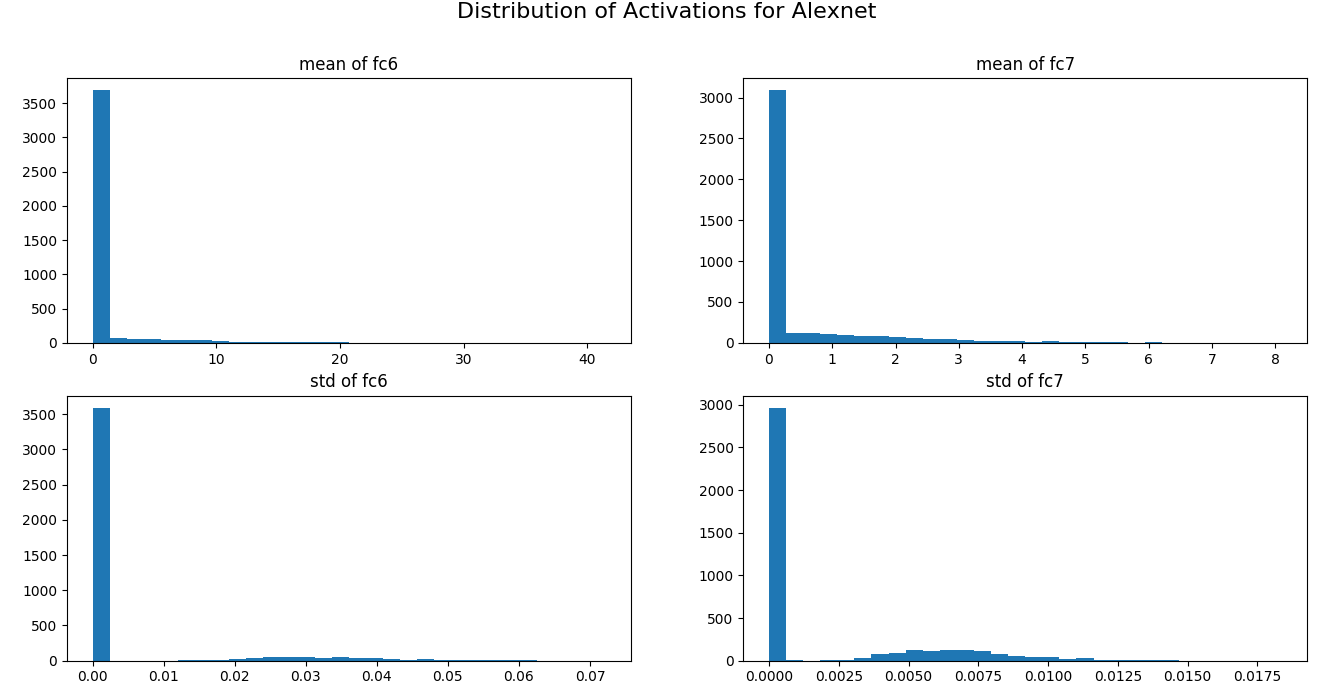}
	\caption{Distributions of mean and standard deviation for fully-connected layers in AlexNet.}
	\label{alexnet_act}
\end{figure}

\begin{figure}[htb]
	\centering
	\includegraphics[width=\columnwidth]{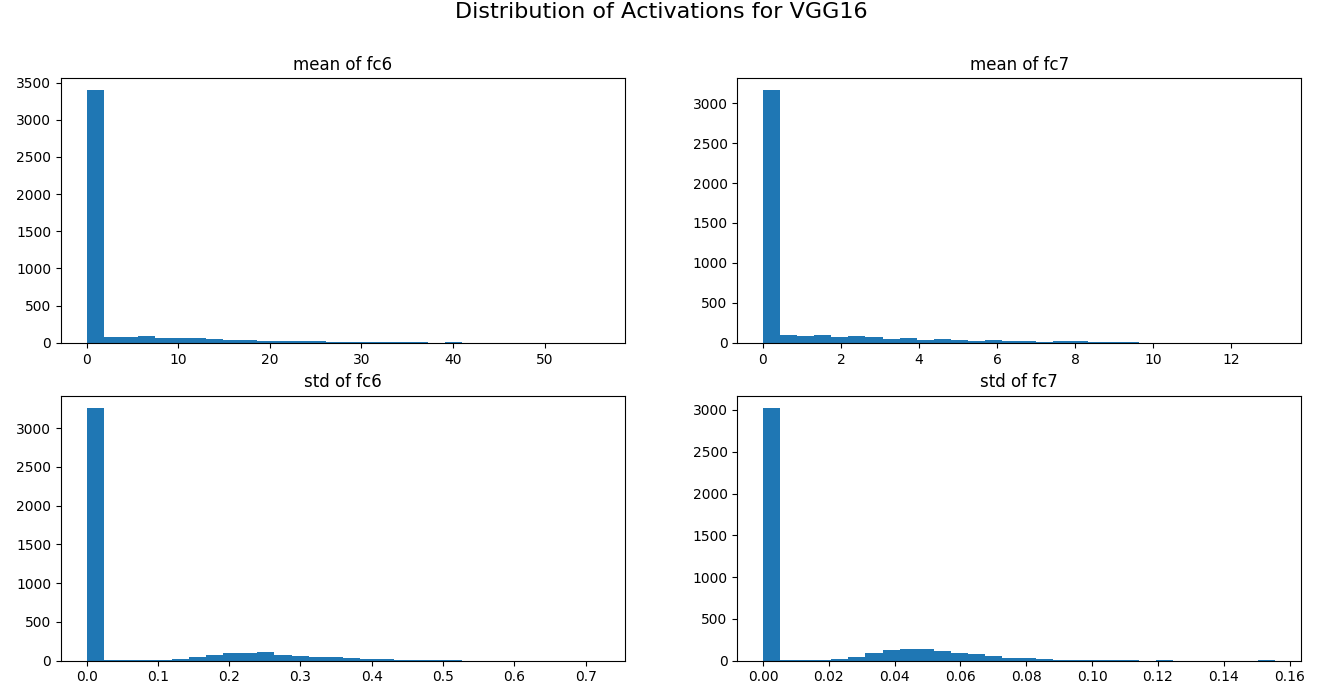}
	\caption{Distributions of mean and standard deviation for fully-connected layers in VGG16.}
	\label{vgg16_act}
\end{figure}

We first choose an arbitrary image from ILSVRC dataset. In order to simulate PDA, we fill the patch of size $10 \times 10$ at the image's top left corner with samples drawn from a conditional Gaussian distribution. A total of $160$ samples are used to make a mini-batch. After feeding the batch to a CNN, we collect the outputs of two fully connected layers, which are also inputs to the following ReLU layers. The distributions of their mean and standard deviation are plotted in histograms \ref{alexnet_act} and \ref{vgg16_act}. 
\noindent
In Figures \ref{alexnet_act} and  \ref{vgg16_act} we can observe that most neurons have their mean and standard deviation at around $0$. Also, the ratio between mean and standard deviation is very large, which meets the requirements for Eq.~\ref{eq:relu1} and Eq.~\ref{eq:relu2} to hold.

\textbf{Nonlinear Components: }
The most commonly used non-linear functions in neural networks are sigmoid functions. It was the default activation function before ReLU was introduced, and it is still used to output probabilities for binary classification tasks. To prove that Eq.~\ref{eg:GM} holds for sigmoid functions, we just need to show that sigmoid functions are a special case of softmax functions:
\begin{align}
\sigma \left( x \right) &= \frac{1}{{1 + \exp \left( { - x} \right)}}
 = \frac{{\exp \left( 0 \right)}}{{\exp \left( 0 \right) + \exp \left( { - x} \right)}}
\end{align}

The observations in this sections lead to a new efficient algorithm to perform prediction evidence analysis. The pseudo-code of this method is shown in Algorithm~\ref{efficient_pda}.

\begin{algorithm}[htb]
	\caption{Efficient Prediction Difference Analysis}
	\label{efficient_pda}
	\begin{algorithmic}[1]
		\State $WE =$ zeros($n*n$), $counts =$ zeros($n*n$)
		\For {every patch $x_w$ of size $k \times k$ in $x$}
		\State $x'$ = copy($x$)
		\State define patch ${\hat x_w}$ of size $l \times l$ that contains $x_w$
		\State ${x_w'}$ = conditional mean of ${x_w}$ given ${\hat x_w}\backslash {x_w}$
		\State $P\left( {\left. c \right|x\backslash {x_w}} \right) = P\left( {c|x'} \right)$
		\State $WE[$coordinates of $x_w]$ += ${\log _2}\left( {odds\left( {c|x} \right)} \right) - {\log _2}\left( {odds\left( {c|x\backslash {x_w}} \right)} \right)$
		\State $counts[$coordinates of $x_w]$ += $1$
		\EndFor
		\State \textbf{return} $WE / counts$
	\end{algorithmic}
\end{algorithm}
\noindent
Algorithm~\ref{efficient_pda} has $S$ times fewer forward passes to run than Algorithm~\ref{pda_code}. Thus it is expected to be at least $S$ times faster.

\section{Evaluation \& Results}
\label{sect:eval}

\textbf{Quantitative Experiments:} 
We have shown that by modelling with normalized geometric mean we can propagate the expectation in output space down to the input space in a layerwise manner. However, it remains a question whether normalized geometric mean provides a good approximation to the arithmetic mean. In this section, we will conduct both theoretical analysis and quantitative  experiments to evaluate the relationship between these two kinds of mean and how it would affect the approximation quality.

The inequality of arithmetic mean and geometric mean imply that the geometric mean of non-negative numbers is less than or equal to the corresponding arithmetic mean. This means that geometric mean consistently underestimates arithmetic mean. However, this property does not hold for normalized geometric mean. For simplicity we consider the binary classification case.

We have shown earlier that the normalized geometric mean for sigmoid function is 
\begin{align}
\begin{split}
NGM\left( {\sigma \left( x \right)} \right) &= \frac{{\prod\limits_{{x_i}} {\sigma {{\left( {{x_i}} \right)}^{P\left( {{x_i}} \right)}}} }}{{\prod\limits_{{x_i}} {\sigma {{\left( {{x_i}} \right)}^{P\left( {{x_i}} \right)}}}  + \prod\limits_{{x_i}} {{{\left( {1 - \sigma \left( {{x_i}} \right)} \right)}^{P\left( {{x_i}} \right)}}} }} \\
&= \sigma \left( {\sum\limits_{{x_i}} {P\left( {{x_i}} \right){x_i}} } \right)
\end{split}
\end{align}
and the arithmetic mean is
\begin{align}
AM\left( {\sigma \left( x \right)} \right) = \sum\limits_{{x_i}} {P\left( {{x_i}} \right)} \sigma \left( {{x_i}} \right)
\end{align}
Since sigmoid function is s-shaped, it is convex on $( - \infty ,0]$ and concave on $[0, + \infty)$. By Jensen's inequality, the normalized geometric mean of sigmoid function is less than or equal to the arithmetic mean for $x_i \le 0$, and it is greater than or equal to the arithmetic mean for $x_i \ge 0$.

Moreover, with first order Taylor series expansion,
\begin{align}
\begin{split}
\log GM\left( {\sigma \left( x \right)} \right) &= \log \prod\limits_{{x_i}} {\sigma {{\left( {{x_i}} \right)}^{P\left( {{x_i}} \right)}}} \\
& = \sum\limits_{{x_i}} {P\left( {{x_i}} \right)} \log \sigma \left( {{x_i}} \right)\\
&\approx \sum\limits_{{x_i}} {P\left( {{x_i}} \right)} \left( {\sigma \left( {{x_i}} \right) - 1} \right)\\
& = AM\left( {\sigma \left( x \right)} \right) - 1\\
GM\left( {\sigma \left( x \right)} \right) &\approx \exp \left( {AM\left( {\sigma \left( x \right)} \right) - 1} \right)\\
& \approx AM\left( {\sigma \left( x \right)} \right)
\end{split}
\end{align}
So geometric mean is a first order approximation of arithmetic mean for sigmoid functions. Based on this, we can show that normalized geometric mean is also a first order approximation of the arithmetic mean:
\begin{align}
\begin{split}
NGM\left( {\sigma \left( x \right)} \right) &= \frac{{\prod\limits_{{x_i}} {\sigma {{\left( {{x_i}} \right)}^{P\left( {{x_i}} \right)}}} }}{{\prod\limits_{{x_i}} {\sigma {{\left( {{x_i}} \right)}^{P\left( {{x_i}} \right)}}}  + \prod\limits_{{x_i}} {{{\left( {1 - \sigma \left( {{x_i}} \right)} \right)}^{P\left( {{x_i}} \right)}}} }}\\
& \approx \frac{{AM\left( {\sigma \left( x \right)} \right)}}{{AM\left( {\sigma \left( x \right)} \right) + \left( {1 - AM\left( {\sigma \left( x \right)} \right)} \right)}}\\
&= {AM\left( {\sigma \left( x \right)} \right)}
\end{split}
\end{align}
It is easy to see that the same argument can be generalized to softmax functions. Since normalized geometric mean does not always underestimate arithmetic mean, it is reasonable to use normalized geometric mean for approximation in our case.

\begin{figure}[htb]
	\centering
	\includegraphics[width=1.0\columnwidth]{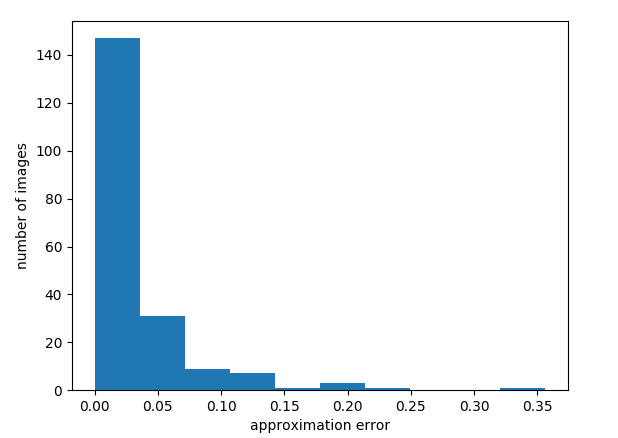}
	\caption{distribution of approximation error for 200 images}
	\label{appro_error}
\end{figure}

In order to visualise the empirical approximation quality, we randomly select a set of $200$ images and a fixed patch location. The arithmetic mean of output probability is approximated by drawing $500$ sample patches at that location, and the normalized geometric mean of output probability is obtained by using conditional mean for that location. The distribution of approximation error is plotted in Figure~\ref{appro_error}. 
\noindent
Figure~\ref{appro_error} shows the differences between arithmetic mean and normalized geometric mean for the output probabilities of AlexNet. We can see that for more than half of the input images the approximation error is very close to zero. In addition to that, as the approximation error increases, fewer images fall into the corresponding intervals. This suggests that our alternative formulation gives a good approximation.
 
\textbf{Computational Speed:}
We run experiments to record the time taken by both algorithms to visualise a single image. In particular, they both use a $10 \times 10$ window size and an $18 \times 18$ outer patch size. $10$ samples are taken for each window location by the original algorithm.

The classifiers used are Alexnet, VGG16 and GoogLenet. The benchmarks are performed using Caffe with CuDNN on a Tesla K80 GPU. A batch size of 160 is used in forward pass.

Note that all the above details have an influence on the overall runtime. However, we are interested int the relative speed-up of our new formulation.
\begin{figure}[htb]
	\centering
	\includegraphics[width=\columnwidth]{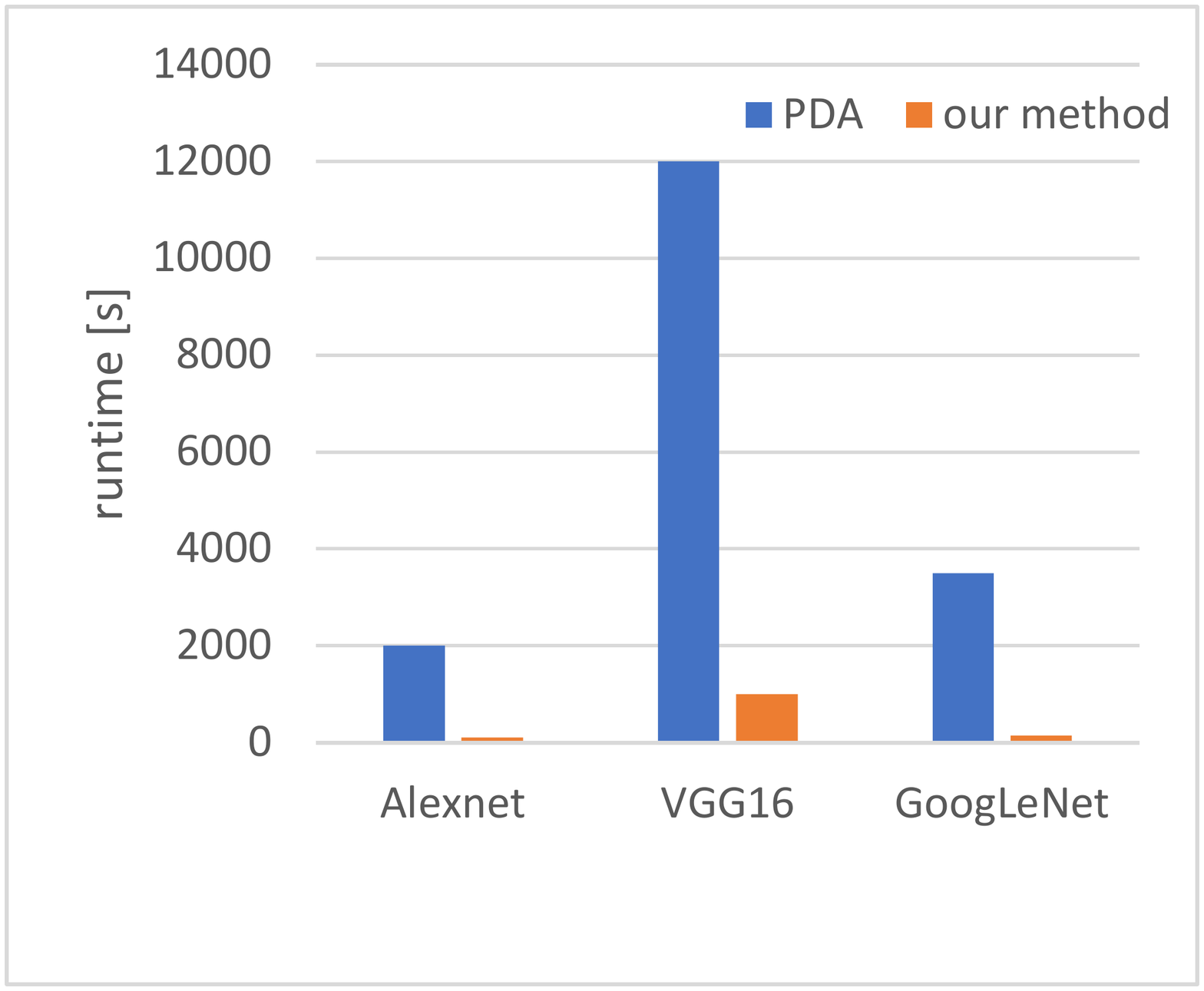}
	\caption{Time required to visualise the decision evidence for a single image by PDA \cite{zintgraf2017visualizing} and our approach.}
	\label{benchmarking}
\end{figure}
\noindent
Figure \ref{benchmarking} summarizes the benchmarking result for our method vs. \cite{zintgraf2017visualizing}. We can see that our proposed modification results in a 10x speed-up, which is a significant improvement.
Furthermore, a batch size of $160$ samples is the maximum for VGG16 to be fit in the GPU memory. For smaller models like Alexnet, we could finish visualisation within minutes by using a much larger batch size.	

\textbf{Qualitative Experiments:}
We train a six-layer convolutional model that consists of two convolutional layers, two max-pooling layers and two fully connected layers. The detailed architecture is illustrated in \cite{deep_mnist}. This model achieves an accuracy of 99.2\% on the MNIST test set.
\begin{figure}[htb]
	\centering
	\includegraphics[width=\columnwidth]{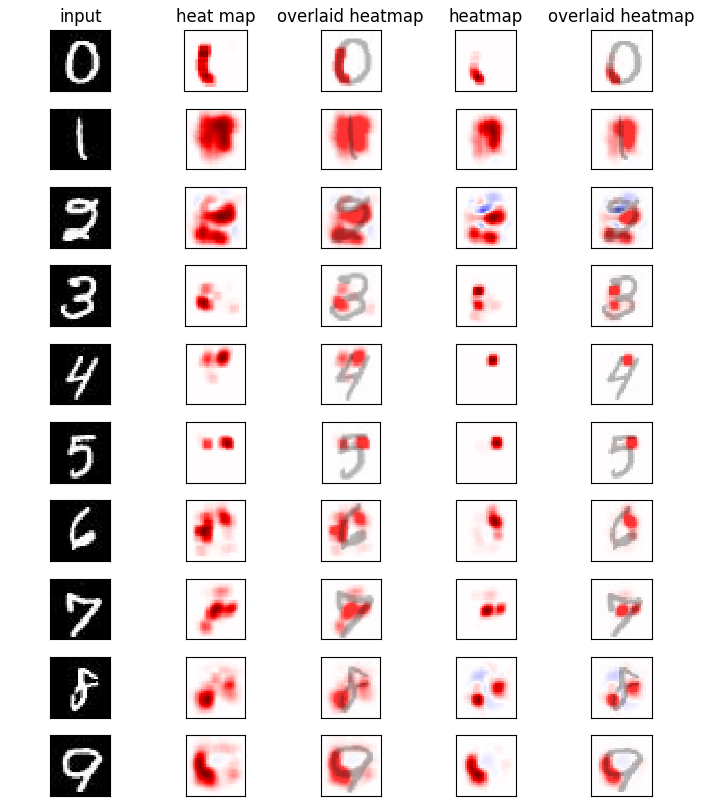}
	\caption{Visualisations for MNIST dataset. The first column shows the input images, which are deliberately chosen to contain digits from zero to nine. The second and fourth columns show the heatmaps generated by \cite{zintgraf2017visualizing} and our approach respectively. The red regions are input pixels supporting the classification decision, while the blue regions are pixels against the decision. The colour intensity is proportional to pixel importance. The third and last columns show the input images overlaid with their heatmaps.}
	\label{mnist_result}
\end{figure}

In Figure \ref{mnist_result} both algorithms use a window size of $4 \times 4$ since it is verified to produce the  smoothest and most interpretable results. Also, marginal sampling is used instead of conditional sampling, which means we now use $ P\left( {x_w} \right)$ to approximate $ P\left( {{x_w}|x\backslash {x_w}} \right)$. This is justified for images from MNIST dataset because their pixels have relatively weak interdependencies.

\textbf{ILSVRC experiment: }
The classifier used for experiments on the ILSVRC 2012 dataset is VGG16, and the parameters of conditional Gaussian distributions are estimated from the validation set of ILSVRC 2012, which contains 50000 images. 

VGG16 is composed of 16 weight layers and 4 max-pooling layers. Since the propagation of expectation we used in deriving our method incurs an approximation error for each layer, we can expect the accumulated error for VGG16 to be larger than the 6-layer model in previous section. In this case, we would like to figure out whether our method could still explain the classifier's decision well.

We use a window size of $10 \times 10$ and outer patch size of $18 \times 18$ for both algorithms. For original prediction difference analysis, we still draw 10 samples for each window location. Since repeated experiments are computationally expensive and our purpose is not to find the optimal set of parameters, we simply adopt the settings from  \cite{zintgraf2017visualizing}.
\begin{figure}[htb]
	\centering
	\includegraphics[width=\columnwidth]{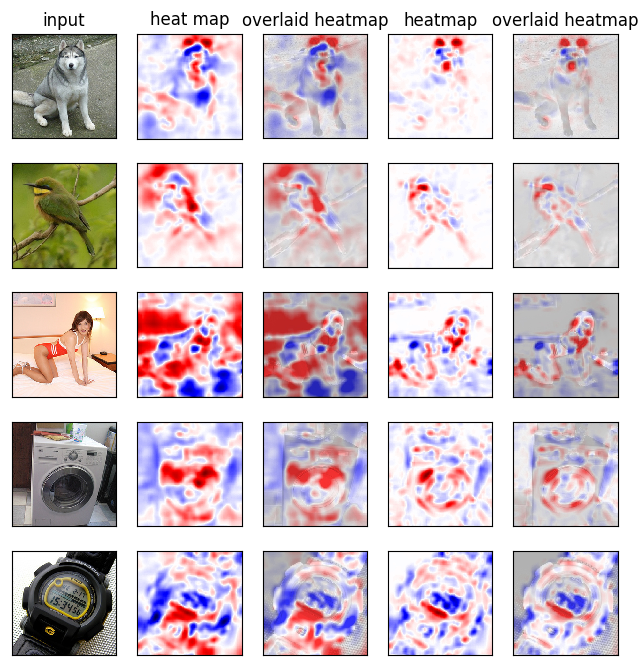}
	\caption{Visualisations for ILSVRC dataset: class labels are "Siberian husky", "bea eater", "maillot", "washer" and "digital watch" respectively in a top-down order. }
	\label{ilsvrc_result1}
\end{figure}

Figure \ref{ilsvrc_result1} shows the visualisation results for five correctly classified images from ILSVRC dataset. 
The column layout is the same as in Figure \ref{mnist_result}.

In the first row of Figure \ref{ilsvrc_result1}, the input image is a Siberian husky. We can observe that both algorithms treat the husky's ears and nose as strong positive evidence while treat its forehead as strong negative evidence. However, the results from \cite{zintgraf2017visualizing} is much noisier; the evidence captured by it spreads over the whole image space, making the heatmap look chaotic. On the other hand, although our method also marks regions outside the object as evidence, their pixel intensities are too low to be confused with the main region. This could help observers to omit unnecessary details and focus on the most important evidence.
The same properties also appear in other examples. In the case of bee eater, both algorithms consider the yellow feather beneath the bird's beak as positive evidence and its tail as negative evidence, but the original algorithm also highlights the region outside the bird's head. In the case of maillot, the original algorithm attributes the highest importance to a large region of wall outside the woman, which is unlikely to help the classifier make its decision.
Furthermore, even inside the objects, the two algorithms can make different judgements. For example, the \cite{zintgraf2017visualizing} decides that the chest region of the husky class votes against this class, but the same region is not considered important by our approach. Also, the two algorithms have contrary evidence assignments for feathers on the bee eater's back.

\begin{figure}[htb]
	\centering
	\includegraphics[width=\columnwidth]{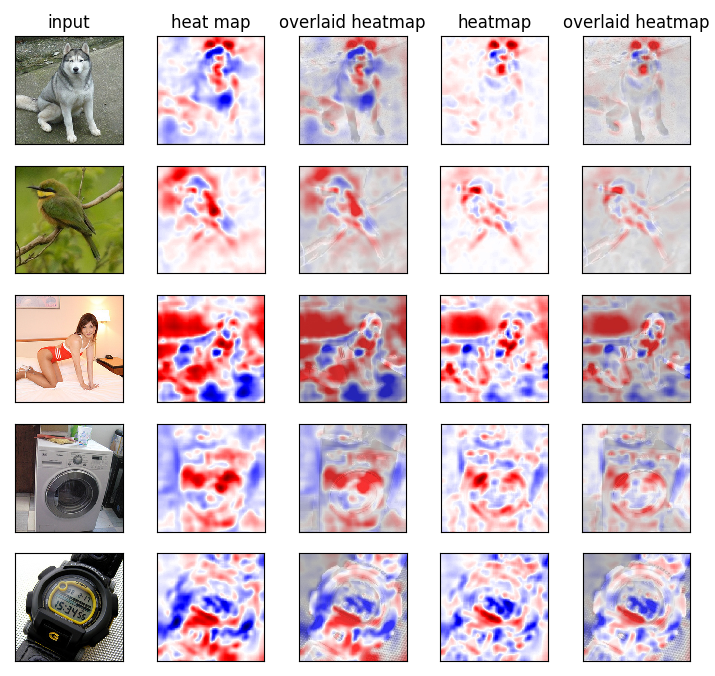}
	\caption{the same as Figure \ref{ilsvrc_result1}, except that visualisations produced by our approach are replaced by their sampled approximations.}
	\label{ilsvrc_result2}
\end{figure}

The above-mentioned differences in feature heatmaps could come from two sources. First, it may be the result of approximation error in our approach. Also, it may be caused by the sampling approach in original prediction difference analysis. In order to differentiate these two kinds of error, we design the following experiment: Intuitively, if we increase the number of samples taken at each window location for original prediction difference analysis, the result would be more accurate and closer to its true value. However, we cannot afford it because it makes this algorithm even more computationally expensive. So we use a different approach to demonstrate the error caused by sampling. Now, we replace the conditional mean in our method with empirical mean of 10 samples instead. The results are shown in Figure~\ref{ilsvrc_result2}.

Figure \ref{ilsvrc_result2} demonstrates the feature heatmaps generated by \cite{zintgraf2017visualizing} and our sampled approach for the same set of input images as \ref{ilsvrc_result1}. 

First of all, we can immediately notice that the regions highlighted by the sampled approximation are nearly the same as those highlighted by original prediction difference analysis for maillot example. In particular, both algorithms agree that the wall is positive evidence, while the bed sheet near the hands is negative evidence. This effect is not so obvious for other examples, whose feature heatmaps look almost identical after switching to empirical mean. However, we can still discover those subtle changes if we take a closer look.

For example, the region above bee eater's head is now taken as positive evidence by the sampled variant of our approach. Also, husky's chest is now taken as negative evidence. These subtleties are difficult to be detected only because their intensities are very low.

Based on these observations, we propose the following hypothesis. For \cite{zintgraf2017visualizing} and our approach, the difference in heatmap's shape is mainly caused by the former's sampling behaviour, while the difference in heatmap's intensity is mainly caused by the latter's approximation error.

To further verify the first part of this hypothesis, we randomly pick an image $x$ and fix a window location $x_w$. We then measure the absolute difference in $P\left( {c|x\backslash {x_w}} \right)$ outputted by the two algorithms when different number of samples are taken at $x_w$. The result is plotted below.
\begin{figure}[htb]
	\centering
	\includegraphics[width=0.6\columnwidth]{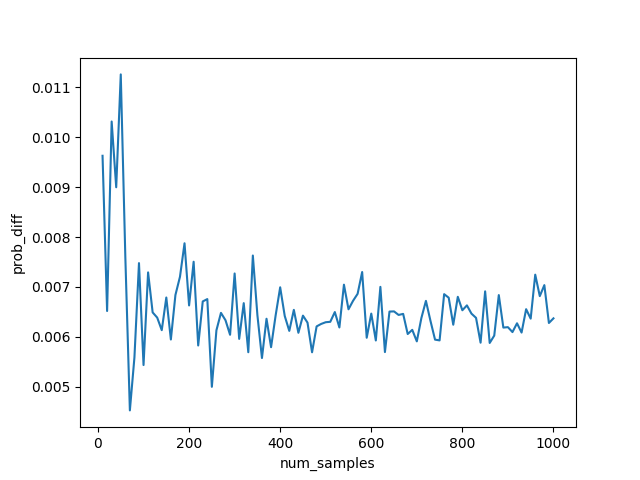}
	\caption{absolute difference in $P\left( {c|x\backslash {x_w}} \right)$ changes with the number of samples drawn at a particular window location $x_w$}
	\label{pred_diff}
\end{figure}
We can see from Figure \ref{pred_diff} that the prediction difference fluctuates intensely when only a few samples are drawn. Since it is infeasible to draw hundreds of samples in practice, it is obvious that \cite{zintgraf2017visualizing} is noisier than our approach.

\textbf{Invariance Experiment:} 
In the previous section, we observe that our approach generates clearer visualisations. However, we cannot affirm that the important regions captured by it are truly discriminative since there is not a general quantitative evaluation of visualisation result. So, what we would show instead is that our method can capture invariant features.

The intuition behind this is that if a CNN generalises well outside its training data, it must have learned a set of invariant features for an object class. If we can show that our method captures similar evidence for different images belonging to the same class, then the evidence is likely to have strong discriminative power.
\begin{figure}[htb]
	\centering
	\includegraphics[width=\columnwidth]{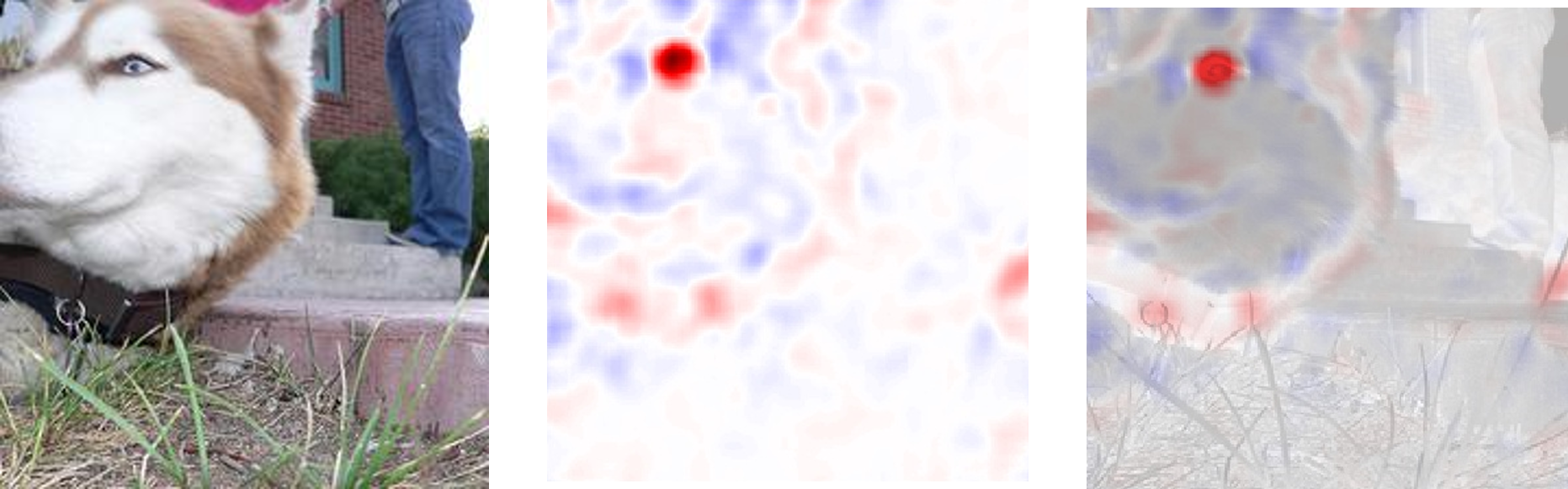}
	\caption{Visualisation result for another husky image.}
	\label{new_husky}
\end{figure}
Figure \ref{new_husky} shows the heatmap generated by our approach for another husky image that is correctly classified by VGG16. It is obvious that the strongest evidence speaking for its class is the husky's eye. If we look back at Figure \ref{ilsvrc_result1}, we would see that the positive evidence found for the first husky image is its ears and nose. Important features found for these two images do not seem to agree with each other.

Before explaining the reason, we shall run further experiments regarding feature invariance. This time, we choose to process the first husky image with operations including rotation, flipping and cropping to see the changes in the evidence analysis.
\begin{figure}[htb]
	\centering
	\includegraphics[width=0.9\columnwidth]{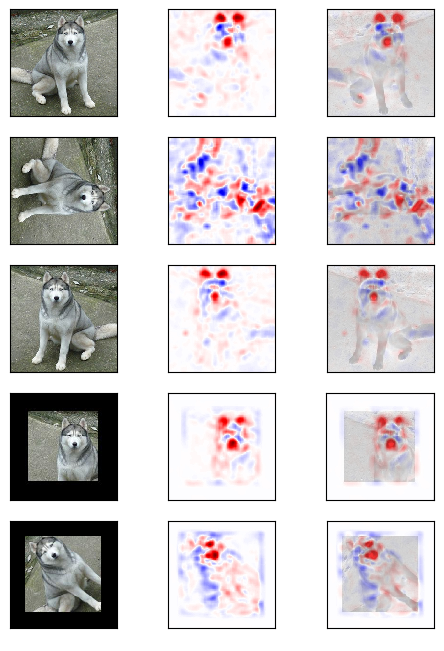}
	\caption{Visualisation results on augmented husky images.}
	\label{huskies}
\end{figure}
In Figure \ref{huskies}, the first row is the original husky image. Row 2 and row 5 are generated by rotation. Row 3 is generated by flipping, and row 4 is generated by cropping.

We can observe that all heatmaps except the last one mark the husky's ears and nose as strong positive evidence. This indicates that the presense of ears and nose can indeed help VGG16 to recognise a husky. As for the last case, when one ear is occluded, VGG16 marks its eye as positive evidence. This behaviour agrees with Figure \ref{new_husky}, in which one ear is occluded as well.

We can even make a guess about VGG16's procedure for recognising huskies. It will first search for the pair of ears and the nose whenever possible. If one ear is missing, then the other one would not be considered important anymore and VGG16 searches for eyes instead.

\textbf{Further Simplification: }
If we ignore log odds, what our approach essentially does for each window $x_w$ is to measure the difference between $P\left( {c|x} \right)$ and $P\left( {c|x'} \right)$, where $x'$ is obtained by replacing $x_w$ with its conditional mean.

Let $f(x)$ be the decision function that outputs $P\left( {c|x} \right)$. Then the first order Taylor expansion of $f(x)$ at $x_0$ is
\begin{align}
f\left( x \right) \approx f\left( {{x_0}} \right) + f'{\left( {{x_0}} \right)^T}\left( {x - {x_0}} \right)
\end{align}
If we let $x_0$ be the original image, we can further approximate our method with the following equation.
\begin{align}
P\left( {c|x} \right) - P\left( {c|x'} \right) \approx f'{\left( x \right)^T}\left( {x - x'} \right)
\end{align}
Hence, we can even further simplify Algorithm~\ref{efficient_pda}. Pseudo-code for the simplified algorithm is presented in Algorithm~\ref{simple_pda}.
\begin{algorithm}[htb]
	\caption{Further simplification of our approach}
	\label{simple_pda}
	\begin{algorithmic}[1]
		\State $WE =$ zeros($n*n$), $counts =$ zeros($n*n$)
		\State $grad = \frac{\partial }{{\partial x}}P\left( {c|x} \right)$
		\For {every patch $x_w$ of size $k \times k$ in $x$}
		\State $x'$ = copy($x$)
		\State define patch ${\hat x_w}$ of size $l \times l$ that contains $x_w$
		\State ${x_w'}$ = conditional mean of ${x_w}$ given ${\hat x_w}\backslash {x_w}$
		\State $WE[$coordinates of $x_w]$ += $grad^T\left( {x - x'} \right)$
		\State $counts[$coordinates of $x_w]$ += $1$
		\EndFor
		\State \textbf{return} $WE / counts$
	\end{algorithmic}
\end{algorithm}
The simplified algorithm now only needs to perform one forward pass and one backward pass to visualise an image. 

To illustrate its visualisation quality, we compare it with our method on ILSVRC dataset. The results are shown below.
\begin{figure}[htb]
	\centering
	\includegraphics[width=\columnwidth]{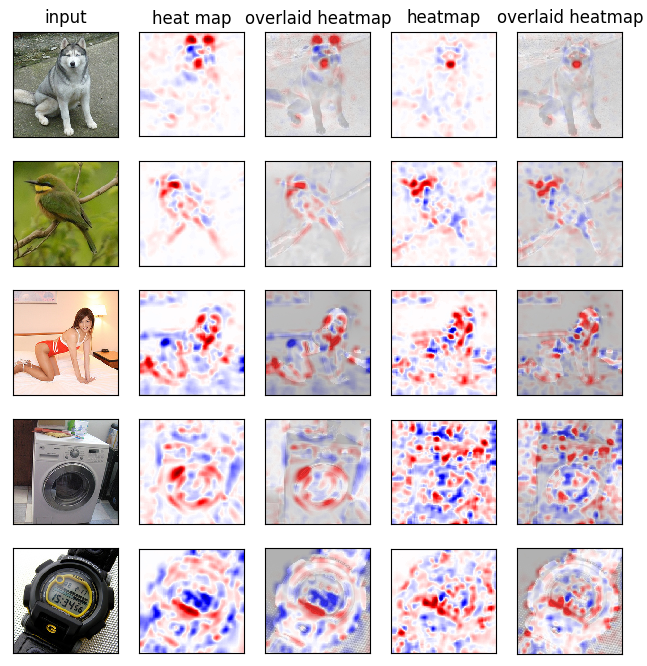}
	\caption{Visualisation results by our approach and its simplified version on the same set of input images as Figure \ref{ilsvrc_result1} and Figure \ref{ilsvrc_result2}}
	\label{ilsvrc_result3}
\end{figure}

In Figure \ref{ilsvrc_result3}, the middle two columns are feature heatmaps generated by our approach, and the last two columns are feature heatmaps generated by its simplified version.

We can see that the simplified version is capable of capturing some important input features, but the results are too noisy and lack interpretability. Although it runs significantly faster, the decrease in visualisation quality would reduce its usefulness.

However, the formulation of simplification itself can help us reason about other visualisation techniques. IN Table~\ref{alg_compare} we compare our method to class saliency map and deep Taylor decomposition in the aspect of feature importance evaluation.

\begin{table}[htb]
	\begin{center}
		\begin{tabular}{ll}
			\toprule
			Visualisation Technique & Importance of Pixel $x_i$ \\ 
			\toprule
			Class Saliency Map~\cite{simonyan2013deep} & ${R_i} = { {\left| {\frac{{\partial f}}{{\partial x_i}}} \right|} }$ \\ 
			\toprule
			Deep Taylor Decomposition~\cite{montavon2017explaining} & ${R_i} = \frac{{\partial f}}{{\partial {{\tilde x}_i}}} \cdot \left( {{x_i} - {{\tilde x}_i}} \right)$ \\ 
			\midrule
			Our approach & ${R_i} = \frac{{\partial f}}{{\partial {x_i}}} \cdot \left( {{x_i} - {x_i'}} \right)$ \\
			\bottomrule
		\end{tabular}
		\caption{three visualisation techniques and their evaluation of pixel importance}
		\label{alg_compare}
	\end{center}
\end{table}
In Table~\ref{alg_compare}, $f$ denotes decision function of the classifier to be visualised. $x$ is the input image, $\tilde x$ is chosen such that $f(\tilde x) = 0$, and $x_i'$ is the conditional mean of $x_i$ given its neighbouring pixels.

We can observe that all three algorithms utilize partial derivatives of the decision function with respect to input pixels to evaluate each pixel's importance. What differentiates them is the weighting rule for pixel contribution. Class saliency maps~\cite{simonyan2013deep} treat every pixel equally, so it only uses gradient information to decide pixel importance. Deep Taylor decomposition~\cite{montavon2017explaining}, on the other hand, first finds a root point of the decision function. It assigns larger weights to pixels far away from this root point because those pixels carry more information about the class identity. Finally, our approach models the interdependence between local pixels and assigns larger weights to pixels that are hard to be predicted from context.

Different weighting rules decide the properties of visualisations produced. Equal weighting is built on the assumption that each pixel independently contributes to the classifier's decision. This is not true for a CNN, which learns a hierarchy of representations from complex pixel interactions. Therefore visualisations generated by class saliency map tend to be noisy. Both Deep Taylor decomposition and our approach weigh pixels proportionally to their distances to a base image; the former's base image encodes information about decision function, while the latter's base image encodes information about distribution of input pixels. Since they explain the classifier's decision from different perspectives, their results can be combined in practice to gain a better interpretation.

\section{Conclusion}

In this work we have investigated a framework for predictor evidence analysis. Our work is based on an alternative formulation for prediction difference analysis~\cite{zintgraf2017visualizing}. 
It is derived by taking advantage of the hierarchical structure and special properties of component functions of CNNs. We have run various experiments on different datasets and classifiers to compare it with other methods aiming to explain classification decisions.
Our proposed method runs at least $10\times$ faster than \cite{zintgraf2017visualizing}. This acceleration comes from reducing the number of forward passes as well as avoiding sampling from high dimensional Gaussian distributions. 
Our method generates more interpretable visualisations. This is due to a side-effect of the new formulation. Instead of using sampling to approximate expectation at output space, the expectation is now taken at input space and we no longer need to draw samples. So, our evidence visualisations are less noisy and the evidence captured focuses more on the objects. 
Our approach can capture invariant class-specific features. If applied to a set of images belonging to the same class, it allows to find the ``decision rules'' of the classifier for recognising that class.
Overall, we also showed that our method can be interpreted as a gradient weighting rule. This interpretation links it with other gradient-based visualisation techniques, which fall into the same mathematical framework.

	{\small
		\bibliographystyle{ieee}
		\bibliography{report}

\begin{thebibliography}{10}\itemsep=-1pt

\bibitem{deep_mnist}
{Deep MNIST for Experts}.
\newblock \url{https://www.tensorflow.org/get_started/mnist/pros}.

\bibitem{bach2015pixel}
S.~Bach, A.~Binder, G.~Montavon, F.~Klauschen, K.-R. M{\"u}ller, and W.~Samek.
\newblock On pixel-wise explanations for non-linear classifier decisions by
  layer-wise relevance propagation.
\newblock {\em PLOS ONE}, 10(7):e0130140, 2015.

\bibitem{baldi2014dropout}
P.~Baldi and P.~Sadowski.
\newblock The dropout learning algorithm.
\newblock {\em Artificial intelligence}, 210:78--122, 2014.

\bibitem{chetlur2014cudnn}
S.~Chetlur, C.~Woolley, P.~Vandermersch, J.~Cohen, J.~Tran, B.~Catanzaro, and
  E.~Shelhamer.
\newblock cudnn: Efficient primitives for deep learning.
\newblock {\em arXiv preprint arXiv:1410.0759}, 2014.

\bibitem{erhan2009visualizing}
D.~Erhan, Y.~Bengio, A.~Courville, and P.~Vincent.
\newblock Visualizing higher-layer features of a deep network.
\newblock {\em University of Montreal}, 1341:3, 2009.

\bibitem{goodfellow2013maxout}
I.~J. Goodfellow, D.~Warde-Farley, M.~Mirza, A.~Courville, and Y.~Bengio.
\newblock Maxout networks.
\newblock {\em arXiv preprint arXiv:1302.4389}, 2013.

\bibitem{jia2014caffe}
Y.~Jia, E.~Shelhamer, J.~Donahue, S.~Karayev, J.~Long, R.~Girshick,
  S.~Guadarrama, and T.~Darrell.
\newblock Caffe: Convolutional architecture for fast feature embedding.
\newblock In {\em Proceedings of the 22nd ACM international conference on
  Multimedia}, pages 675--678. ACM, 2014.

\bibitem{montavon2017explaining}
G.~Montavon, S.~Lapuschkin, A.~Binder, W.~Samek, and K.-R. M{\"u}ller.
\newblock Explaining nonlinear classification decisions with deep taylor
  decomposition.
\newblock {\em Pattern Recognition}, 65:211--222, 2017.

\bibitem{nguyen2016synthesizing}
A.~Nguyen, A.~Dosovitskiy, J.~Yosinski, T.~Brox, and J.~Clune.
\newblock Synthesizing the preferred inputs for neurons in neural networks via
  deep generator networks.
\newblock {\em NIPS 2016}, 2016.

\bibitem{nguyen2015deep}
A.~Nguyen, J.~Yosinski, and J.~Clune.
\newblock Deep neural networks are easily fooled: High confidence predictions
  for unrecognizable images.
\newblock In {\em Proceedings of the IEEE Conference on Computer Vision and
  Pattern Recognition}, pages 427--436, 2015.

\bibitem{rabinovich2007objects}
A.~Rabinovich, A.~Vedaldi, C.~Galleguillos, E.~Wiewiora, and S.~Belongie.
\newblock Objects in context.
\newblock In {\em Computer vision, 2007. ICCV 2007. IEEE 11th international
  conference on}, pages 1--8. IEEE, 2007.

\bibitem{ruderman1994statistics}
D.~L. Ruderman.
\newblock The statistics of natural images.
\newblock {\em Network: computation in neural systems}, 5(4):517--548, 1994.

\bibitem{selvaraju2016grad}
R.~R. Selvaraju, A.~Das, R.~Vedantam, M.~Cogswell, D.~Parikh, and D.~Batra.
\newblock Grad-cam: Why did you say that? visual explanations from deep
  networks via gradient-based localization.
\newblock {\em arXiv preprint arXiv:1610.02391}, 2016.

\bibitem{simonyan2013deep}
K.~Simonyan, A.~Vedaldi, and A.~Zisserman.
\newblock Deep inside convolutional networks: Visualising image classification
  models and saliency maps.
\newblock {\em arXiv preprint arXiv:1312.6034}, 2013.

\bibitem{yosinski2015understanding}
J.~Yosinski, J.~Clune, A.~Nguyen, T.~Fuchs, and H.~Lipson.
\newblock Understanding neural networks through deep visualization.
\newblock In {\em ICML Deep Learning Workshop}, 2015.

\bibitem{zeiler2014visualizing}
M.~D. Zeiler and R.~Fergus.
\newblock Visualizing and understanding convolutional networks.
\newblock In {\em European conference on computer vision}, pages 818--833.
  Springer International Publishing, 2014.

\bibitem{zeiler2011adaptive}
M.~D. Zeiler, G.~W. Taylor, and R.~Fergus.
\newblock Adaptive deconvolutional networks for mid and high level feature
  learning.
\newblock In {\em Computer Vision (ICCV), 2011 IEEE International Conference
  on}, pages 2018--2025. IEEE, 2011.

\bibitem{zhou2016learning}
B.~Zhou, A.~Khosla, A.~Lapedriza, A.~Oliva, and A.~Torralba.
\newblock Learning deep features for discriminative localization.
\newblock In {\em Proceedings of the IEEE Conference on Computer Vision and
  Pattern Recognition}, pages 2921--2929, 2016.

\bibitem{zintgraf2017visualizing}
L.~M. Zintgraf, T.~S. Cohen, T.~Adel, and M.~Welling.
\newblock Visualizing deep neural network decisions: Prediction difference
  analysis.
\newblock {\em arXiv preprint arXiv:1702.04595}, 2017.

\end{thebibliography}
	}
	
\end{document}